\documentclass[letterpaper, 10 pt, journal, twoside]{IEEEtran}

\usepackage[T1]{fontenc}

\usepackage[space, compress, sort]{cite}
\usepackage{url}
\usepackage[linkcolor=black, citecolor=black, urlcolor=black, colorlinks=true]{hyperref}
\usepackage[top=20.1mm, bottom=15.2mm, left=16.9mm, right=16.9mm]{geometry}

\usepackage{amssymb, amsfonts, amsthm, mathtools, bm}

\usepackage[ruled,vlined]{algorithm2e}
\usepackage{algpseudocode}
\SetKwRepeat{Do}{do}{while}
\usepackage[dvipsnames,table]{xcolor}
\usepackage{graphicx}
\usepackage{float}
\usepackage{tabularx, array, multirow, booktabs, makecell}

\usepackage{enumerate}
\usepackage{verbatim}
\usepackage[normalem]{ulem}
\usepackage{colortbl}
\usepackage{pifont}

\usepackage{pifont}
\definecolor{forestgreen}{RGB}{34,139,34}
\definecolor{firebrick}{RGB}{178,34,34}
\newcommand{\cmark}{\textcolor{forestgreen}{\ding{51}}}
\newcommand{\xmark}{\textcolor{firebrick}{\ding{55}}}
\usepackage{enumitem}
\usepackage{subfigure}

\usepackage{booktabs,multirow,array}
\usepackage[table]{xcolor}     
\usepackage{siunitx}
\definecolor{ourshl}{gray}{0.92}   
\newcolumntype{P}{S[table-format=3.1]}  %
\newcolumntype{E}{S[table-format=1.3]}  %
\newcolumntype{L}{S[table-format=3.2]}  %
\newcolumntype{A}{S[table-format=1.2]}  %
\newcolumntype{V}{S[table-format=2.2]}  %
\usepackage{tabularx}
\usepackage{makecell}
\usepackage{threeparttable}
\newcolumntype{Y}{>{\raggedright\arraybackslash}X}
\newcommand{\upmetric}{\raisebox{0.12ex}{\scriptsize$\uparrow$}}
\newcommand{\downmetric}{\raisebox{0.12ex}{\scriptsize$\downarrow$}}

\newcommand{\delete}[1]{\bgroup\markoverwith{\textcolor{red}{\rule[0.5ex]{2pt}{0.4pt}}}\ULon{#1}}
\definecolor{lightgreen}{RGB}{220, 255, 220}
\definecolor{lightyellow}{RGB}{255, 255, 200}
\definecolor{lightred}{RGB}{255, 220, 220}
\definecolor{headerblue}{RGB}{200, 220, 255}
\newcommand{\Method}{EFLUX\xspace}


\title{ \LARGE \bf EFLUX: Elastic Multi-Robot Formation Navigation and Adaptation with Agentic LLMs
}

\author{Jinyuan Zhang\textsuperscript{1}, 
Yuwei Wu\textsuperscript{1}, 
Guangyao Shi\textsuperscript{2}, 
Jonathan Diller\textsuperscript{1}, 
Gaurav S. Sukhatme\textsuperscript{2}, 
Vijay Kumar\textsuperscript{1}
\thanks{\textsuperscript{1}Jinyuan Zhang, Yuwei Wu, Jonathan Diller, and Vijay Kumar
are with the GRASP Lab, University of Pennsylvania, Philadelphia, PA 19104, USA. 
Email: \texttt{\{jinyuanz, yuweiwu, diller, kumar\}@seas.upenn.edu}.
\textsuperscript{2}Guangyao Shi and Gaurav S. Sukhatme are with the Department of Computer Science, 
University of Southern California, Los Angeles, CA 90089, USA. 
Email: \texttt{\{shig, gaurav\}@usc.edu}.
We greatly acknowledge the support of TILOS, funded by NSF Grant CCR-2112665.
}}

\begin{document}

\maketitle
\thispagestyle{empty}

\begin{abstract}

Multi-robot teams operating in confined or cluttered environments must adapt both their formation geometry and group topology to navigate through complex obstacles. 
This adaptation requires two complementary behaviors: \emph{deformation}, where the team continuously reshapes its geometry while remaining connected, and \emph{reconfiguration}, where robots split into subgroups or merge back into a single formation.
Existing methods often model these behaviors independently, connect them through handcrafted rules, or lack explicit geometric criteria for determining when each behavior should be invoked.
However, challenging environments may require online changes in formation shape, connectivity, and effective team composition, making decoupled or rule-based approaches prone to suboptimal trajectories and deadlock.
We propose \Method, a geometry-grounded LLM agentic framework for automatic and elastic multi-robot formation navigation.
\Method extracts a structured scene representation and uses an LLM to reason jointly over both deformation actions, such as scaling and shearing, and reconfiguration actions, such as splitting and merging. 
These strategies are then translated into executable per-robot waypoints through a closed-loop generation, verification, and correction pipeline. 
Simulation and hardware experiments show that \Method enables safe, continuous, and elastic formation navigation in constrained environments, reducing deadlock and navigation failures compared with baselines while maintaining coherent multi-robot coordination.

\end{abstract}

\section{Introduction}

Multi-robot formation navigation is a fundamental capability for collaborative tasks such as search and rescue~\cite{hu2021decentralized}, cooperative transportation~\cite{alonso2017multi}, and exploration~\cite{zhou2022swarm}. 
In cluttered environments, a robot team must reach its target positions and avoid collisions while maintaining desired formation~\cite{7487747}, yet these objectives often conflict. 
Rigid formation maintenance may cause deadlock in narrow corridors~\cite{desai2002modeling} or choke points~\cite{9812172}, whereas prioritizing collision avoidance alone leads to excessive formation distortion and loss of coordination~\cite{9812050}. 
Robust formation navigation therefore requires adaptive mechanisms that modify team geometry in response to environmental constraints while preserving coordinated behaviors.

\begin{figure}[!t]
      \centering
    \includegraphics[width=1\columnwidth]{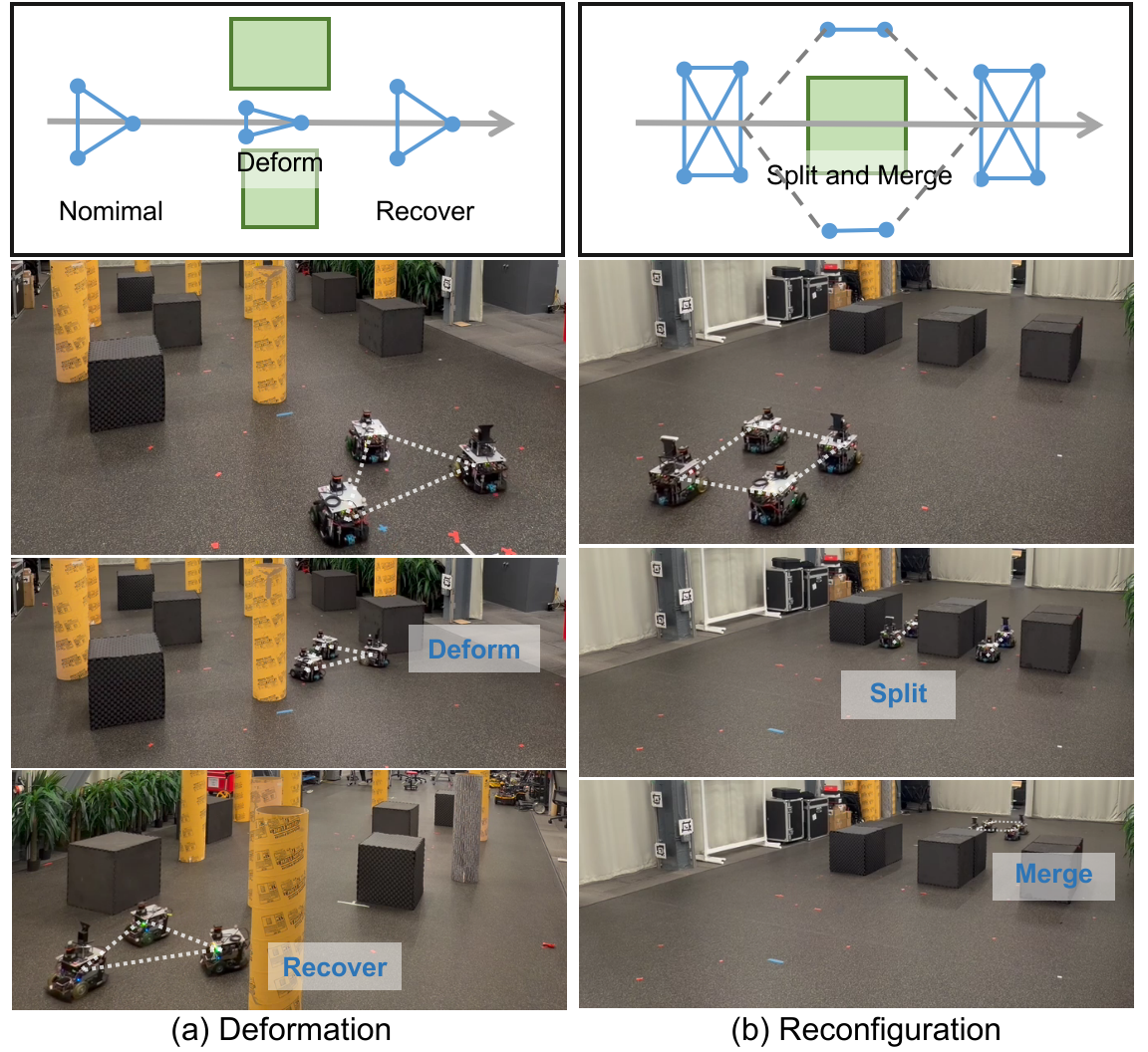}
    \vspace{-0.6cm}
      \caption{Formation (a) deformation and (b) reconfiguration of a robot team.
      }
      \label{fig:fig1}
      \vspace{-0.2cm}
\end{figure}

Existing methods approach formation navigation from control, optimization, and learning perspectives, but they adapt formations only in limited ways: classical control and optimization-based methods rely on fixed structures, predefined templates, or handcrafted switching logic~\cite{lewis1997high,desai2002modeling,8429106,ren2005consensus,quan2023robust,alonso2017multi,zhou2022swarm,7487747,zhou2025number,li2025deform}, while learning-based policies are usually reactive and lack explicit high-level reasoning about when and how to adapt~\cite{yan2022relative,deng2025coordinated}. 
Effective formation adaptation requires both \textit{deformation}, the continuous modification of formation geometry, and \textit{reconfiguration}, discrete topological changes corresponding to split and merge behaviors. 
These two adaptations are usually complementary: deformation alone may preserve connectivity but fail in narrow passages that cannot accommodate the full robot team, whereas reconfiguration alone may over-fragment the team into too many small groups and increase coordination complexity. 
Thus, moderate constraints can often be handled by deforming a connected formation, while severe bottlenecks require temporary reconfiguration into subgroups that traverse separately before merging back into the nominal formation (Fig.~\ref{fig:fig1}).
The major challenge is not to execute a predefined formation change, but to \textbf{determine automatically when to preserve, deform, or reconfigure the formation during navigation.} 
Prior work, however, addresses these behaviors separately and thus lacks a unified framework for continuous geometric and topological adaptation.

Recent advances in large language models (LLMs) provide a promising direction for unified, joint reasoning over formation adaptation.
LLMs have demonstrated strong capabilities in semantic interpretation, contextual reasoning, and code generation. 
In multi-robot formation tasks, they are mainly used for static pattern generation~\cite{liu2024language}, shape assembly~\cite{zhu2025lamarl}, or automated reward design~\cite{yao2025application}, rather than for reasoning about formation adaptation during navigation. 
This motivates using LLMs as high-level reasoning modules that interpret scene structure and select adaptation strategies. 
However, LLM reasoning alone lacks the geometric precision required for safe formation navigation, as its direct outputs are probabilistic and prone to hallucination, often producing infeasible plans. 
Executable formation adaptation requires \textbf{geometric grounding}: high-level decision making must be constrained by scene representations, spatial constraints, and feasibility checks before being translated into robot actions.
Existing LLM-based methods lack such explicit deterministic geometric grounding and cannot reliably ensure feasible formation adaptation.

To address this challenge, we propose \textbf{EFLUX} (\textbf{E}lastic \textbf{F}ormations with \textbf{L}LMs), an LLM-based agentic framework for multi-robot formation navigation in cluttered environments. EFLUX combines deterministic geometric grounding with language-driven reasoning to jointly handle formation deformation and reconfiguration. 
It first extracts a structured scene representation from the map information, user-specified nominal formation, and task specification, then interprets this representation to localize geometric bottleneck contexts. It then generates adaptation strategies and converts them into verified waypoints through a closed-loop generation-and-verification pipeline before execution by low-level controllers.
EFLUX is evaluated in both simulation and hardware experiments across diverse scenarios requiring deformation, reconfiguration, and combined adaptation.
Our contributions are as follows:
\begin{itemize}
\item We propose a unified, geometry-grounded LLM framework for automatic formation navigation that jointly handles elastic adaptation in team composition and formation shape through reconfiguration and deformation.

\item We ensure geometric grounding through deterministic geometric abstraction and a closed-loop reasoning pipeline that validates and refines LLM-proposed plans for collision-free, safely executable navigation.

\item We validate the framework in simulation and hardware, demonstrating higher success rates and fewer failures than baselines, with generalization across team sizes, formation shapes, and dynamic membership changes.
\end{itemize}

\begin{table}[t]
  \centering
  \caption{Feature comparison for multi-robot formation coordination.}
  \label{tab:feature_comparison}
  \scriptsize
  \setlength{\tabcolsep}{2.5pt}
  \renewcommand{\arraystretch}{1.06}
  \begin{threeparttable}
  \begin{tabular*}{\columnwidth}{@{\extracolsep{\fill}}lccccc@{}}
    \toprule
    \textbf{Methods}
    & \makecell{Deformation}
    & \makecell{Reconfigu-\\ration}
    & \makecell{Dynamic\\Membership}
    & \makecell{Shape\\Adaptive}
    & \makecell{Core\\Method} \\
    \midrule

    VRB~\cite{8429106}
    & \cmark & \xmark & \xmark & \xmark & VRB+APF \\

    STAF~\cite{dengsubteaming}
    & \cmark & \cmark & \xmark & \xmark & GNN+RL \\

    DEFORM~\cite{li2025deform}
    & \cmark & \xmark & \xmark & \cmark & MPC \\

    DVS~\cite{zhou2025number}
    & \cmark & \xmark & \cmark & \xmark & MINCO \\

    SF~\cite{9812050}
    & \cmark & \xmark & \xmark & \xmark & MINCO \\

    Rel-Form~\cite{xie2025multi}
    & \cmark & \xmark & \xmark & \xmark & MAPPO \\

    LGPF~\cite{liu2024language}
    & \xmark & \xmark & \xmark & \cmark & LLM+MADDPG \\

    \midrule
    \textbf{EFLUX} (ours)
    & \cmark & \cmark & \cmark & \cmark & LLM+MPC \\

    \bottomrule
  \end{tabular*}

  \begin{tablenotes}[flushleft]
    \scriptsize
    \item *Support: \cmark yes, \xmark no.
Deformation: Continuous geometric reshaping of a connected formation.
Reconfiguration: Splitting the team into subgroups and merging back.
Dynamic Membership:  Allowing robots to join or leave (e.g., upon robot failure or new robot joining) while the coordinated formation task continues without restarting or deadlock.
Shape Adaptive: Switching among predefined formation shapes during the task.
  \end{tablenotes}
  \end{threeparttable}
  \vspace{-0.1em}
\end{table}

\section{Related Works}
\noindent \textbf{Multi-Robot Formation Navigation.}
Prior formation navigation falls into control-, optimization-, and learning-based approaches (Table~\ref{tab:feature_comparison}).
Classical methods specify desired geometries through virtual structures~\cite{lewis1997high} and leader--follower ~\cite{desai2002modeling}, often combined with artificial potential fields~\cite{8429106} or consensus~\cite{ren2005consensus} for distributed motion.
While effective in open environments, these methods assume static formations and lack active adaptation, making swarms prone to deadlock in cluttered scenes~\cite{9812050}.
Affine transformations~\cite{zhao2018affine} relax this by allowing shearing and scaling, yet such continuous deformation alone cannot handle discrete topological changes such as team splitting.
These control-theoretic approaches offer strong guarantees but remain task-specific.
Optimization and learning-based methods improve adaptability through collision-aware trajectory optimization~\cite{alonso2017multi,park2023formation}, deformable formations that adapt geometry from sensed traversability~\cite{li2025deform}, and decentralized policy learning with obstacle avoidance, curriculum learning, and reconfiguration coordination~\cite{yan2022relative,yuan2025multiagent,xie2025multi,dengsubteaming}.
Despite these advances, existing methods typically adapt reactively or rely on predefined templates and rules, limiting their ability to provide all capabilities summarized in Table~\ref{tab:feature_comparison}.

\noindent \textbf{LLMs for Multi-Robot Coordination.} 
LLMs have recently been applied to robot motion planning and control through natural-language interaction~\cite{brohan2023can}, code generation~\cite{liang2022code}, and task planning~\cite{huang2022language}, and extended to multi-robot coordination by code-as-policy pipelines such as LAN2CB~\cite{huang2025compositional} and GenSwarm~\cite{ji2026genswarm}, which convert mission descriptions into runnable multi-robot code.
For formation control and navigation specifically, LGPF~\cite{liu2024language} translates language prompts into static patterns realized by RL policies, while LAMARL~\cite{zhu2025lamarl} and related work~\cite{yao2025application} use LLMs to generate prior policies or shape rewards. Yet these remain limited to one-shot pattern generation or offline training, and to our knowledge none performs closed-loop formation adaptation during navigation.

\begin{figure*}[!t]
 \vspace{0.1cm}
      \centering
    \includegraphics[width=2\columnwidth]{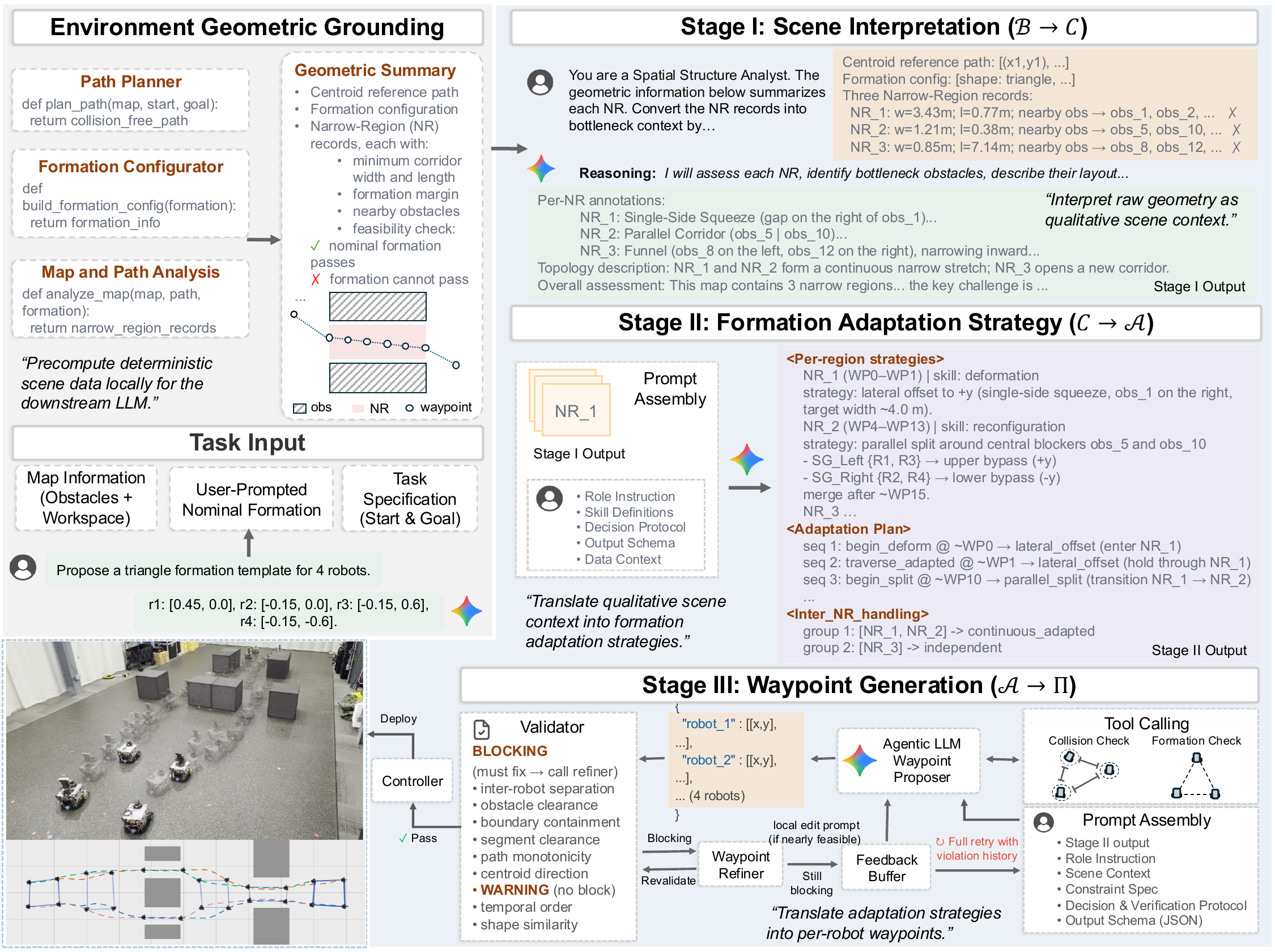}
    \vspace{-0.2cm}
      \caption{
        System overview of \textbf{\Method}. Given a map, task specification, and a user-specified nominal formation, the framework uses deterministic geometric grounding to identify narrow regions that require formation adaptation. It then applies a three-stage agentic LLM pipeline consisting of scene interpretation, formation adaptation strategy generation, and closed-loop waypoint generation, producing verified per-robot waypoints for execution by a low-level controller.
      }
      \label{fig:overview}
      \vspace{-0.5cm}
\end{figure*}

\section{Problem Formulation}

Consider \(N\) disk-shaped robots of radius \(r>0\) navigating in a workspace \(\mathcal{X}\subset\mathbb{R}^2\) with known static convex obstacles \(\mathcal{O}\) and boundary \(\partial\mathcal{X}\). Let \(\delta_{\mathrm{safe}}>0\) denote a safety margin.
Following~\cite{quan2023robust}, the formation is modeled as a weighted undirected graph \(\mathcal{G}=(\mathcal{V},\mathcal{E})\), where \(\mathcal{V}=\{1,\dots,N\}\) indexes the robots, \(\mathcal{E}\subseteq\{\{i,j\}: i,j\in\mathcal{V},\, i\neq j\}\), and edge weights encode squared inter-robot distances. Let \(\mathbf{A}\in\mathbb{R}^{N\times N}\) denote the adjacency matrix with \(A_{ij}=\|\mathbf{p}_i-\mathbf{p}_j\|^2\) for \(\{i,j\}\in\mathcal{E}\), and let \(\mathbf{L}=\mathbf{D}-\mathbf{A}\) be the graph Laplacian with degree matrix \(\mathbf{D}\). The normalized Laplacian is defined as $\hat{\mathbf{L}}=\mathbf{D}^{-1/2}\mathbf{L}\mathbf{D}^{-1/2} $.
Let $\mathcal{F}=\{\mathbf{p}_i^{\mathrm{nom}}\}_{i=1}^{N}$ denote a fixed nominal formation template with topology $\mathcal{E}_{\mathrm{nom}}$. Let $\hat{\mathbf{L}}_{\mathrm{nom}}$ be its normalized Laplacian, which serves as a translation-, rotation-, and uniform-scale-invariant descriptor of the formation geometry. During navigation, the planner generates a time-varying desired formation \(\mathcal{F}^{\mathrm{des}}(t)\) through two modes: \textit{deformation} and \textit{reconfiguration}. 

Over a time interval \([t_a,t_b]\), deformation is modeled as a time-varying weighted graph
\(\mathcal{G}^{\mathrm{des}}(t)=(\mathcal{V},\mathcal{E}^{\mathrm{des}}(t))\)
with fixed topology and continuous edge weights, characterized by the normalized Laplacian
\(\hat{\mathbf{L}}^{\mathrm{des}}(t)\):
\begin{enumerate}[label=(\roman*)]
    \item \textit{Fixed membership and topology:} $\mathcal{V}$ is fixed and
$\mathcal{E}^{\mathrm{des}}(t)\equiv\mathcal{E}_{\mathrm{nom}}$ on $[t_a,t_b]$.
    
    \item \textit{Continuous deformation:} \(A_{ij}^{\mathrm{des}}(t)=\|\mathbf{p}_i^{\mathrm{des}}(t)-\mathbf{p}_j^{\mathrm{des}}(t)\|^2\) varies continuously for all \(\{i,j\}\in\mathcal{E}^{\mathrm{des}}(t)\). 
    
    \item \textit{Connectivity preservation:} \(\lambda_2(\hat{\mathbf{L}}^{\mathrm{des}}(t))>0\) holds for all \(t\in[t_a,t_b]\).
\end{enumerate}
A representative example is an affine deformation, in which the nominal formation template is continuously transformed through linear operations such as shearing, and compression~\cite{zhao2018affine}.
Reconfiguration introduces discrete topological changes by partitioning
the vertex set $\mathcal{V}$ into independently evolving connected subgroups.
$\mathcal{G}^{\mathrm{des}}(t)
=
\bigsqcup_{k=1}^{K(t)}
\mathcal{G}_k^{\mathrm{des}}(t)$,
with the \(k\)-th subgroup graph $
\mathcal{G}_k^{\mathrm{des}}(t)
= (\mathcal{V}_k(t),\mathcal{E}_k^{\mathrm{des}}(t))
$. The following conditions hold:
\begin{enumerate}[label=(\roman*)]
    \item \textit{Fixed membership:}
    \(
    \mathcal{V}
    =
    \bigsqcup_{k=1}^{K(t)}
    \mathcal{V}_k(t)
    \)
    remains unchanged.
    
    \item \textit{Subgroup connectivity:}
    \(
    \lambda_2(\hat{\mathbf{L}}_k^{\mathrm{des}}(t))>0
    \)
    for
    \(
    |\mathcal{V}_k(t)|\ge2
    \).

    \item \textit{Topological variation:}
    the partition
    \(
    \{\mathcal{V}_k(t)\}_{k=1}^{K(t)}
    \)
    changes at least once within
    \(
    (t_a,t_b)
    \),
    corresponding to split or merge events.
\end{enumerate}

A representative example is splitting the team into parallel subgroups to traverse multiple corridors before merging back into a connected formation.

\textbf{Problem:} We consider multi-robot formation navigation in a cluttered environment. 
Given the problem specification \(\mathcal{I}=(\mathcal{X},\mathcal{O},\mathcal{X}_s,\mathcal{X}_g,\mathcal{F})\), a robot team must navigate from the start \(\mathcal{X}_s\) to the goal \(\mathcal{X}_g\) while preserving the nominal formation \(\mathcal{F}\) wherever the free space can accommodate it. Here, \(\mathcal{X}_s,\mathcal{X}_g\in\mathcal{X}\) denote the start and goal positions of the team centroid. In regions where the nominal formation cannot be safely maintained, the planner generates time-varying adapted formations through deformation or reconfiguration.
EFLUX aims to generate collision-free synchronized trajectories while minimizing formation distortion and traversal costs, subject to dynamic and safety constraints.
During reconfiguration, the team temporarily splits into subgroups before merging back into a connected formation. 
In open space, nominal formation preservation is prioritized, whereas in constrained areas, safety and navigational feasibility dominate.

\section{Elastic Multi-Robot Formations with LLMs}

We introduce an elastic multi-robot formation navigation pipeline (Fig.~\ref{fig:overview}) that integrates deterministic geometric planning with geometry-grounded LLM-guided adaptation in constrained environments.
Deterministic modules compute collision-free centroid paths, extract narrow-region geometry, and verify safety, while the LLM interprets geometric bottlenecks and generates deformation or reconfiguration strategies with synchronized waypoint plans. 

To couple geometric grounding with semantic reasoning, we introduce narrow-region records $\mathcal{B}$, encoding traversability and local obstacle structure along the centroid path $\gamma$. As shown in Algorithm~\ref{alg:planner_flow}, the framework follows the representation chain $\mathcal{B}\rightarrow\mathcal{C}\rightarrow\mathcal{A}\rightarrow\Pi$, where the LLM interprets the records $\mathcal{B}$ as a semantic scene context $\mathcal{C}$, selects adaptation strategies $\mathcal{A}$, and proposes waypoint plans $\Pi$. Candidate plans are then validated deterministically, with failed plans producing structured feedback $\mathcal{H}$ for iterative replanning.

\begin{algorithm}[!th]
    \footnotesize
    \LinesNumbered
    \setlength{\algomargin}{1.1em}
    \SetAlCapFnt{\footnotesize}
    \SetAlCapNameFnt{\footnotesize}
    \caption{Formation planning with narrow-region records}
    \label{alg:planner_flow}

    \KwIn{Information $\mathcal{I}$, planner parameters $\Theta$, maximum attempt number $K$}
    \KwOut{Admissible plan $\Pi$; on failure, the best candidate $\Pi^\star$
           over the $K$ attempts}
    \BlankLine

    $(\gamma,\mathcal{B})\leftarrow$ EnvironmentAnalysis$(\mathcal{I},\Theta)$\;
    Construct the deterministic validator $\mathcal{M}$ from $\mathcal{I}$ and $\Theta$\;
    $\mathcal{C}\leftarrow$ SceneAnalysis$(\mathcal{B},\gamma,\mathcal{O})$\;
    $\mathcal{A}\leftarrow$ AdaptationStrategy$(\mathcal{C},\gamma,\mathcal{F},\Theta)$\;
    Initialize $\Pi^\star\leftarrow\emptyset$, $v^\star\leftarrow+\infty$,
    $\mathcal{H}\leftarrow\emptyset$
    \tcp*[r]{$v^\star$: best blocking-violation score so far}
    \For{$k=1,\ldots,K$}{
      $\Pi_k\leftarrow$ WaypointGeneration$(\mathcal{A},\mathcal{C},\gamma,\mathcal{F},\Pi^\star,\mathcal{H})$\;
      $(\tilde{\Pi}_k,v_k)\leftarrow$ ValidateAndRefine$(\Pi_k,\mathcal{M})$
      \tcp*[r]{validate; if violations remain, run refiner and revalidate}
      \If{$v_k$ has no blocking violation}{
          \Return{$\tilde{\Pi}_k$}\;
      }
      \If{$v_k$ has fewer blocking violations than $v^\star$}{
          $(\Pi^\star,v^\star)\leftarrow(\tilde{\Pi}_k,v_k)$\;
      }
      $\rho_k\leftarrow$ \textsc{RepairMode}$(v^\star)\in\{\textsc{local-fix},\textsc{full-replan}\}$\
      \tcp*[r]{choose next-retry mode}
      Append $(v_k,\rho_k)$ to $\mathcal{H}$ \tcp*[r]{log this attempt's feedback}
    }
    \Return{Failure with $\Pi^\star$}\;
\end{algorithm}

\subsection{Deterministic Environment Analysis}
\label{sec:env}

Rather than passing the raw occupancy map to LLMs, this module converts the workspace into abstract narrow-region records with geometric summaries.
We first obtain a coarse centroid reference path $\gamma$ using a sampling-based planner that adaptively densifies samples near turns.
The planner enforces single-robot feasibility by requiring every point on $\gamma$ to maintain at least $r+\delta_{\mathrm{safe}}$ clearance from obstacles.
We then upsample the path $\gamma$ at a finer resolution to obtain measurement waypoints $\{\mathbf{q}_m\}_{m=0}^{M}$, at which we evaluate the corridor width $w_m^{\mathrm{cor}}$ and formation margin $c_m^{\mathrm{form}}$. The corridor width $w_m^{\mathrm{cor}}$ is estimated by casting $N_{\mathrm{ray}}$ opposing ray pairs from $\mathbf{q}_m$ over $360^\circ$ and taking the minimum summed first-hit distance across all pairs.
This approximates the narrowest free-space cross-section through $\mathbf{q}_m$, independent of formation geometry.
To evaluate whether the nominal formation can traverse the path without deformation or reconfiguration, we place $\mathcal{F}$ with its centroid at each $\mathbf{q}_m$ and its orientation aligned with the local path direction.
Let $\mathbf{p}_i(\mathbf{q}_m)$ denote the resulting position of robot $i$.
The formation margin is defined as
\begin{equation}
c_m^{\mathrm{form}}
= \min_i \left[ \min_{o\in\mathcal{O}\cup\partial\mathcal{X}}
\mathrm{dist}\bigl(\mathbf{p}_i(\mathbf{q}_m),o\bigr) -r
\right]
\end{equation}
representing the minimum robot-to-obstacle or robot-to-boundary clearance of the nominal formation at $\mathbf{q}_m$.
Consecutive waypoints that the nominal formation cannot clear ($c_m^{\mathrm{form}}<\delta_{\mathrm{safe}}$) are grouped into discrete narrow-region $\mathrm{NR}_j$, each summarizing its minimum corridor width, length, and bounding obstacles to form $\mathcal{B}$.
\subsection{Hierarchical LLM Reasoning}
\label{sec:llm}

\textbf{Scene Interpretation ($\mathcal{B}\rightarrow\mathcal{C}$).} \label{sec:stage1}
The records $\mathcal{B}$ quantify the geometric severity of each narrow region $\mathrm{NR}_j$, but do not explain the obstacle configuration causing the bottleneck or the spatial relationships among adjacent regions along $\gamma$. Both are important for deciding whether deformation or reconfiguration is required. To provide this context, the LLM augments each $\mathrm{NR}_j$ with a severity label, obstacle-grounded spatial description, nominal-formation passability assessment, and inter-region topological relations, to produce the semantic scene context $\mathcal{C}$.

The LLM is conditioned on the obstacle layout, reference path, and geometric summaries in $\mathcal{B}$, with bottleneck-forming obstacles separated from nearby contextual obstacles. Since narrow regions are identified upstream, the LLM reasons over one bounded neighborhood at a time rather than the entire workspace. A guided protocol prompts it to identify the local obstacle layout, classify the bottleneck pattern (e.g., funnel, corridor, chicane, or single-side squeeze), and perform a reverse-consistency check before populating the schema. All geometric quantities are deterministically computed, with the LLM restricted to qualitative interpretation rather than geometric estimation.

\noindent \textbf{Formation Adaptation Strategy ($\mathcal{C}\rightarrow\mathcal{A}$).}
\label{sec:stage2}
This stage performs high-level adaptation selection and scheduling. For each $\mathrm{NR}_j$, the LLM determines whether the team should remain connected through deformation or temporarily reconfigure into subgroups, and schedules these adaptations along $\gamma$ to avoid unnecessary adapt--recover oscillations. To reduce the LLM's numerical reasoning burden, we inject precomputed geometric summaries to provide the LLM with a geometry-grounded heuristic decision rule. Let $w_j=\min_{\mathbf{q}_m\in \mathrm{NR}_j} w_m^{\mathrm{cor}}$ denote the minimum width of $\mathrm{NR}_j$, and let $W_{\mathrm{nom}}$ denote the path-lateral footprint width of the nominal formation. The width ratio $\rho_j=w_j/W_{\mathrm{nom}}$ measures the severity of each narrow region relative to the nominal formation, while the safety-reduced admissible width $w_j^{\mathrm{safe}}=w_j-2\delta_{\mathrm{safe}}$ defines the maximum formation footprint width allowed. Inter-region gap lengths classify neighboring narrow regions as independent, continuous, or progressive. These quantities are combined with the Stage I semantic annotations, reference path, and obstacles.

The LLM follows a guided Chain-of-Thought (CoT) protocol with reverse-prediction checks that simulate the resulting team behavior against the Stage I scene description before generating the strategy. The decision process defaults to deformation to preserve the nominal formation descriptor $\hat{\mathbf{L}}_{\mathrm{nom}}$ whenever feasible. A connected deformation is width-admissible when $w_j^{\mathrm{safe}}\ge2r$. If this condition fails, connected deformation through the same passage is rejected, and the planner considers reconfiguration where feasible subgroup traversal channels exist. Reconfiguration is also preferred when the scene contains multiple traversable corridors or obstacle geometries that cannot be handled effectively through affine deformation alone. Even when deformation is geometrically feasible, reconfiguration may still be selected if the required distortion would significantly violate formation-shape preservation. To guide the LLM at this decision boundary, the prompt provides canonical pattern-strategy mappings: a narrowing funnel suggests progressive scaling, and a dual-bypass obstacle suggests parallel subgroup reconfiguration with robots grouped by the nearest corridor.

The generated adaptation plan specifies both strategy type and execution schedule. For deformation, it specifies the deformation mode (e.g., scaling, shearing, or line formation), compression axis, target width, and lateral bias relative to $\gamma$, which together parameterize the desired formation shape.  For reconfiguration, it specifies split type (parallel, sequential, or hybrid), subgroup assignments with traversal channels, merge locations, and optional subgroup-local deformation. The LLM additionally records a CoT trace containing path-level reasoning, region grouping, strategy evaluation, and feasibility checks. 
A final validation step then verifies that the plan covers all narrow regions and is internally consistent before it is passed to Stage III.

\begin{table}[!th]
  \centering
  \footnotesize
  \caption{Stage III tool-calling and deterministic validator checks.}
  \label{tab:tools}
  \setlength{\tabcolsep}{2.6pt}
  \renewcommand{\arraystretch}{1.1}
  \begin{tabularx}{\linewidth}{@{}l l >{\raggedright\arraybackslash}X@{}}
    \toprule
    Interface & Input & Check \\
    \midrule

    \multicolumn{3}{@{}l}{\textit{Tool-calling checks}} \\

    Point & $\mathbf{p}$ &
    $d_{\mathcal{O}}(\mathbf{p}) \ge r+\delta_{\mathrm{safe}}$
    and
    $d_{\partial\mathcal{X}}(\mathbf{p}) \ge r+\delta_{\mathrm{safe}}$ \\

    Segment/path & $(\mathbf{p}_a,\mathbf{p}_b)$ or $\tau_i$ &
    $\min_{\zeta\in[0,1]}
    d_{\mathcal{O}}(\operatorname{seg}_{ab}(\zeta))
    \ge r+\delta_{\mathrm{safe}}$,
    applied to one segment or all consecutive segments in $\tau_i$ \\

    Snapshot & $\{\mathbf{p}_i(t)\}_{i=1}^{N}$ &
    $\|\mathbf{p}_i(t)-\mathbf{p}_j(t)\|
    \ge 2(r+\delta_{\mathrm{safe}})$ for $i\neq j$;
    compute the realized normalized formation Laplacian \(\hat{\mathbf{L}}(t)\) \\

    \midrule

    \multicolumn{3}{@{}l}{\textit{Deterministic validator checks}} \\

    Schema & $\Pi$ &
    JSON validity, robot IDs, synchronized waypoint counts, and valid waypoint states \\

    Plan & $\Pi$ &
    Blocking checks for inter-robot separation, obstacle clearance,
    boundary containment, segment clearance, path monotonicity, and
    centroid-direction consistency; temporal order, formation-shape error,
    and smoothness are retained as diagnostics \\

   \bottomrule
    \addlinespace[2pt]

    \multicolumn{3}{@{}>{\raggedright\arraybackslash}p{\linewidth}@{}}{%
      \scriptsize
      $d_{\mathcal{O}}$ and $d_{\partial\mathcal{X}}$ denote obstacle and
      boundary distances, respectively, while
      $\operatorname{seg}_{ab}(\zeta)=(1-\zeta)\mathbf{p}_a+\zeta\mathbf{p}_b$
      denotes a queried segment.%
    }\\
  \end{tabularx}
\end{table}

\noindent \textbf{Waypoint Generation ($\mathcal{A}\rightarrow\Pi$).} \label{sec:stage3}
This stage performs coordinate-level instantiation by converting the selected adaptation schedule into synchronized waypoint sequences along $\gamma$.
The LLM is conditioned on scene, formation, path, and constraint context, and is prompted to produce a strict JSON waypoint plan satisfying safety, synchronization, directional consistency, and temporal ordering.

Geometric validity is enforced outside the LLM. 
Stage III treats the LLM as a heuristic waypoint proposer within a tool-calling and validation loop: each candidate plan is checked by the interfaces shown in Table~\ref{tab:tools}, locally refined when nearly feasible, and otherwise fed back to the LLM through structured violation records.
The deterministic validator $\mathcal{M}$ applies plan-level checks to the complete waypoint trajectory and returns a verdict $v_k$. Blocking violations reject the candidate, while non-blocking diagnostics are retained for feedback-guided retry.
Thus, $\mathcal{M}$ enforces admissibility and consistency rather than trajectory optimality.

\subsection{Agentic Validation, Retry, and Replanning}
\label{sec:retry}

Discrete formation strategies may produce waypoint plans $\Pi_k$ that violate geometric constraints such as obstacle clearance.
To improve robustness, Stage III is augmented with an agentic correction loop: the LLM proposes candidate plans, while deterministic modules validate, refine, and provide structured feedback for subsequent retries.
Before re-querying the LLM, a constrained local refiner attempts to repair small coordinate-level violations:
\begin{equation}
\begin{aligned}
\tilde{\Pi}_k=\arg\min_{\Pi,\epsilon}\quad&
\mathcal{J}_{\mathrm{track}}(\Pi,\Pi_k)
+\lambda_{\mathrm{rep}}\mathcal{J}_{\mathrm{rep}}(\Pi)
+\lambda_{\epsilon}\|\epsilon\|_2^2\\
\mathrm{s.t.}\quad&
\mathbf{g}(\Pi,\epsilon)\ge0,\qquad
\|\Pi-\Pi_k\|_{\infty}\le\delta_{\max},
\end{aligned}
\end{equation}
where 
\(\mathbf{g}:=\{g_{\mathrm{sep}},g_{\mathrm{obs}},g_{\mathrm{seg}}\}\) enforces inter-robot separation and waypoint/segment obstacle clearance, $\Pi_k$ denotes the LLM-generated waypoint plan and $\Pi$ the refined waypoint coordinates. 
The objective preserves fidelity to the proposed plan while repelling waypoints from nearby obstacles; slack variables $\epsilon$ are penalized to maintain feasibility. 
The trust-region bound $\delta_{\max}>0$ restricts the refinement to local geometric corrections without altering the formation topology.

The refined candidate is accepted only if it does not increase the blocking-violation score after revalidation; otherwise, the original proposal is retained for feedback and retry. Remaining violations are encoded as structured feedback records containing the violated constraint type, associated robots or obstacles, waypoint or segment indices, and quantitative clearance errors. These records are accumulated in a feedback buffer $\mathcal{H}$ and appended to subsequent prompts during the retry loop with attempt budget $K$ (Algorithm~\ref{alg:planner_flow}). Across retries, $\mathcal{C}$ and $\mathcal{A}$ remain fixed, while waypoint generation is conditioned on $\gamma$, $\mathcal{F}$, the best candidate $\Pi^\star$, and $\mathcal{H}$. When a nearly feasible solution exists, the prompt explicitly requests minimal edits to violating waypoints rather than full regeneration. 

The same framework also supports online replanning. 
Upon an external trigger, the controller pauses execution and constructs a replanning snapshot from the stopped robot states, remaining reference path $\gamma_{\mathrm{rem}}$, and unexecuted waypoint suffix $\Pi_{\mathrm{rem}}$. 
Mission-level changes (e.g., goal updates) trigger full replanning from the current team centroid, while team-level changes (e.g., robot failures or additions) reuse $\gamma_{\mathrm{rem}}$ and update the team configuration or formation template.
The resulting plan is subjected to the same validation, retry, and transition checks before execution, enabling fault-tolerant and dynamically adaptive formations.

\section{Validation Experiments}

We conduct simulation and hardware experiments to evaluate EFLUX. The experiments validate three claims: (1) EFLUX outperforms representative baselines in constrained formation-navigation tasks by jointly reasoning over deformation and reconfiguration; (2) geometric grounding over both the robot team and environment, together with deterministic validation and retry, improves the reliability of LLM-generated waypoints; and (3) EFLUX is elastic in both formation shape and team size, supporting reconfiguration, scaling to different numbers of robots, and dynamic membership changes caused by robot failures or additions. We first describe the implementation details, then benchmark EFLUX against baselines across diverse constrained scenarios, analyze stage-wise ablations, LLM backbones, and scalability, and finally validate the framework on hardware platforms.

\subsection{Implementation Details and Metrics}

We use Gemini-3.1-Pro~\cite{deepmind2026gemini3} as the LLM backbone for all planning stages, with the temperature set to 0.2 to balance deterministic behavior and limited exploration. Stage~III allows up to five retries, and the candidate waypoint refiner is solved by CasADi~\cite{Andersson2019}. The accepted multi-robot waypoints are then time-parameterized into synchronized reference trajectories and executed in Gazebo simulation using a distributed MPC waypoint tracker. Each robot is modeled as a differential-drive platform with unicycle kinematics. The tracker runs at 10~Hz with a 12-step prediction horizon, prioritizing waypoint tracking and control smoothness, together with a short-range obstacle avoidance term that provides redundancy against residual tracking errors. The maximum linear speed is set to 2.2~m/s, and the maximum angular speed is set to 2~rad/s.

The following metrics are used to evaluate performance.
$\mathrm{SR}$ denotes the task success rate. $T_{\mathrm{plan}}$ and $T_{\mathrm{tr}}$ denote the planning and traversal times in seconds, respectively. $N_{\mathrm{att}}$ is the average number of Stage~III waypoint-generation attempts, while $V_{\text{1st}}$ counts the blocking violations in the first Stage~III proposal. $N_{\mathrm{tok}}$ denotes the total number of LLM tokens used during planning. 
\(e_{\mathrm{sim}}\) denotes the average normalized-Laplacian based formation similarity error, computed over the full robot set as the squared Frobenius norm \(\|\hat{\mathbf{L}}(t)-\hat{\mathbf{L}}_{\mathrm{nom}}\|_F^2\), treating the swarm as a single formation at each time step~\cite{quan2023robust}. $L_{\mathrm{total}}$ denotes the summed path length of all robots.

\begin{table}[t]
    \centering
    \caption{Comparison across five constrained scenarios.}
    \label{tab:main_comparison}
    \setlength{\tabcolsep}{2pt}
    \renewcommand{\arraystretch}{1.0}
    \scriptsize
    \resizebox{\columnwidth}{!}{%
    \begin{tabular}{@{} l l cccc cccc @{}}
    \toprule
    \multirow{2}{*}{\textbf{Scenario}} & \multirow{2}{*}{\textbf{Method}}
    & \multicolumn{4}{c}{\textbf{Triangle formation}}
    & \multicolumn{4}{c}{\textbf{Square formation}} \\
    \cmidrule(lr){3-6} \cmidrule(lr){7-10}
    & & SR\,(\%)\,$\uparrow$ & $T_{\text{tr}}$\,(s)\,$\downarrow$
        & $e_{\text{sim}}$\,$\downarrow$ & $L_{\text{total}}$\,(m)\,$\downarrow$
      & SR\,(\%)\,$\uparrow$ & $T_{\text{tr}}$\,(s)\,$\downarrow$
        & $e_{\text{sim}}$\,$\downarrow$ & $L_{\text{total}}$\,(m)\,$\downarrow$ \\
    \midrule
    \multirow{4}{*}{\shortstack[l]{Narrow corridor\\(gap $=1.2$\,m)}}
      & DEFORM         &   95.0 &   92.6 &  0.217 &  42.55 &   70.0 &  120.8 &  0.167 &  44.28 \\
      & VRB            &  \textbf{100.0} &   \textbf{51.6} &  0.048 &  40.17 &  \textbf{100.0} &   \textbf{50.5} &  0.045 &  40.05 \\
      & SF             &  \textbf{100.0} &   75.0 &  \textbf{0.005} &  41.31 &  \textbf{100.0} &   71.6 &  \textbf{0.011} &  40.68 \\
      & Ours           &  \textbf{100.0} &   57.8 &  0.013 &  \textbf{38.79} &  \textbf{100.0} &   57.6 &  \textbf{0.011} &  \textbf{38.71} \\
    \midrule
    \multirow{4}{*}{\shortstack[l]{Narrow corridor\\(gap $=0.7$\,m)}}
      & DEFORM         &   95.0 &  170.4 &  0.162 &  44.59 &   85.0 &  176.4 &  0.147 &  45.64 \\
      & VRB            &    0.0 &     -- &     -- &     -- &    0.0 &     -- &     -- &     -- \\
      & SF             &   70.0 &   \textbf{54.0} &  \textbf{0.053} &  41.01 &   90.0 &   \textbf{53.8} &  0.043 &  41.08 \\
      & Ours           &  \textbf{100.0} &   63.6 &  0.084 &  \textbf{39.21} &  \textbf{100.0} &   59.3 &  \textbf{0.021} &  \textbf{38.98} \\
    \midrule
    \multirow{4}{*}{\shortstack[l]{Single large\\obstacle}}
      & DEFORM         &   45.0 &  122.6 &  0.189 &  45.31 &    0.0 &     -- &     -- &     -- \\
      & VRB            &    0.0 &     -- &     -- &     -- &    0.0 &     -- &     -- &     -- \\
      & SF             &    0.0 &     -- &     -- &     -- &   20.0 &  113.1 &  0.047 &  63.52 \\
      & Ours           &  \textbf{100.0} &   \textbf{67.6} &  \textbf{0.001} &  \textbf{43.94} &  \textbf{100.0} &   \textbf{68.8} &  \textbf{0.000} &  \textbf{43.83} \\
    \midrule
    \multirow{4}{*}{\shortstack[l]{Obstacle\\environment \\ (sparse)}}
      & DEFORM         &  \textbf{100.0} &  188.4 &  0.278 &  44.14 &   10.0 &  238.9 &  0.279 &  49.94 \\
      & VRB            &  \textbf{100.0} &   \textbf{59.5} &  0.054 &  42.81 &      50.0 &  \textbf{62.0} &  0.042 &  43.62 \\
      & SF             &  \textbf{100.0} &   71.0 &  0.048 &  47.32 &   90.0 &   62.3 &  0.023 &  44.09 \\
      & Ours           &  \textbf{100.0} &   79.5 &  \textbf{0.040} &  \textbf{42.55} &   \textbf{95.0} &   80.3 &  \textbf{0.020} &  \textbf{41.72} \\
    \midrule
    \multirow{4}{*}{\shortstack[l]{Obstacle\\environment \\ (dense)}}
      & DEFORM         &    0.0 &     -- &     -- &     -- &    1.2 &  300.9 &  0.275 & 103.16 \\
      & VRB            &    0.0 &     -- &     -- &     -- &   40.0 &  \textbf{131.9} &  0.095 &  99.22 \\
      & SF             &   47.5 &  181.4 &  \textbf{0.040} & 112.44 &   68.8 &  175.9 &  \textbf{0.039} & 111.17 \\
      & Ours           &   \textbf{53.8} &  \textbf{176.6} &  0.087 &  \textbf{98.19} &   \textbf{88.8} &  173.6 &  0.087 &  \textbf{97.31} \\
    \bottomrule
    \multicolumn{10}{p{10.5cm}}{*Best result per column is in \textbf{bold}; ``--''
      marks a method with no successful run. Each entry uses 20 trials; Dense is evaluated with 4 maps $\times$ 20 trials $= 80$. $T_{\text{tr}}$,
      $e_{\text{sim}}$, and $L_{\text{total}}$ are averaged over successful runs. All methods are evaluated under identical start-goal settings and obstacle layouts using both triangle and square formations.
} \\
    \end{tabular}%
    }
\end{table}

\subsection{Benchmark Evaluation in Diverse Scenarios}

We compare EFLUX against three representative baselines: DEFORM~\cite{li2025deform}, which switches formation patterns according to estimated traversable width and tracks them using distributed MPC; VRB~\cite{8429106}, which employs a virtual rigid-body structure with potential-field-based collision avoidance; and SF~\cite{9812050}, which formulates formation navigation as trajectory optimization with obstacle and formation-similarity costs.

\begin{figure}
    \vspace{0.1cm}
    \centering
    \subfigure[Traversal through a 0.7 m-gap corridor.]{
      \includegraphics[width=0.98\columnwidth]{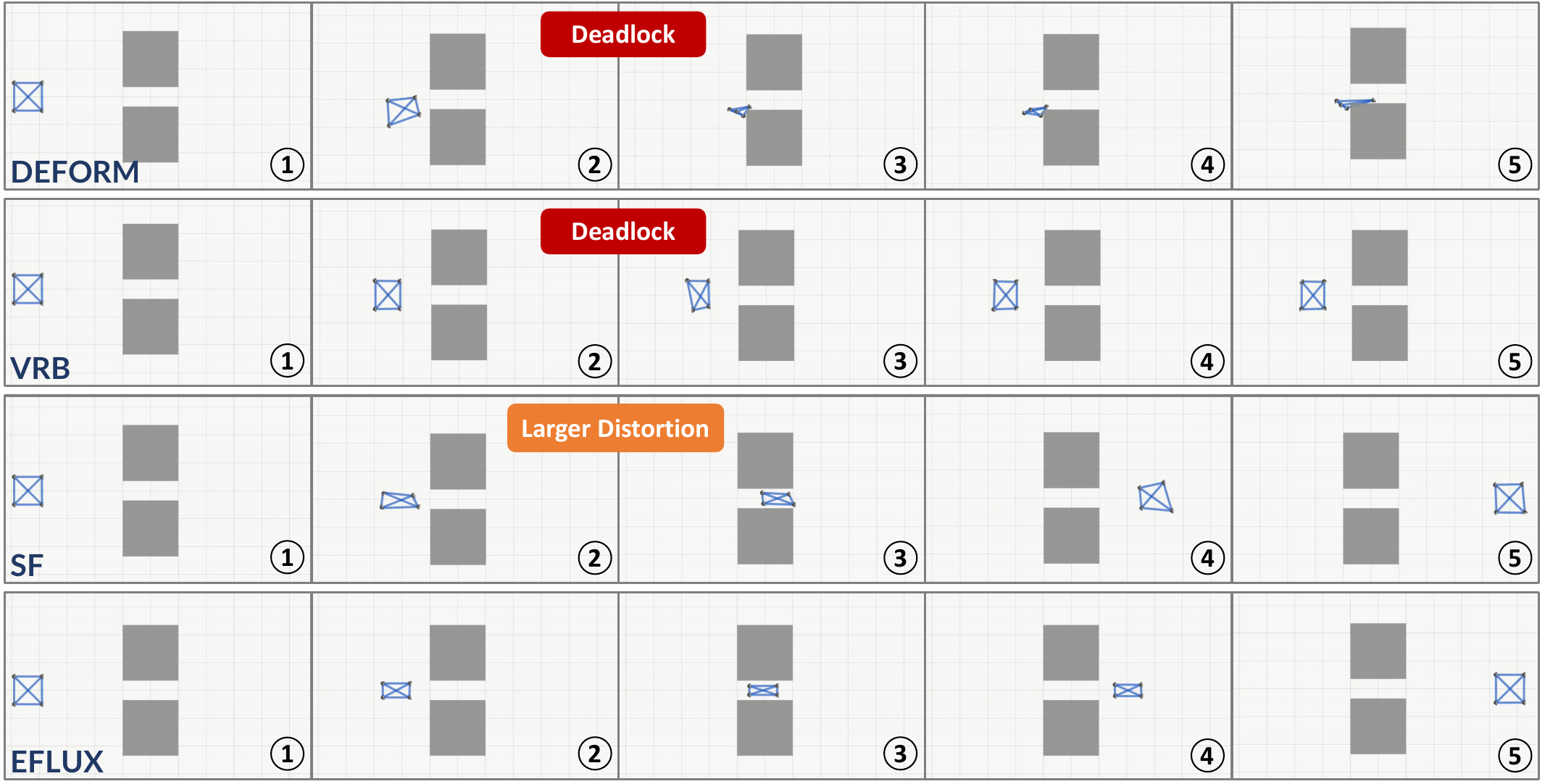}
      \label{fig:model_comparison_corridor}
    }
    \subfigure[Traversal through an obstacle environment.]{
      \includegraphics[width=0.98\columnwidth]{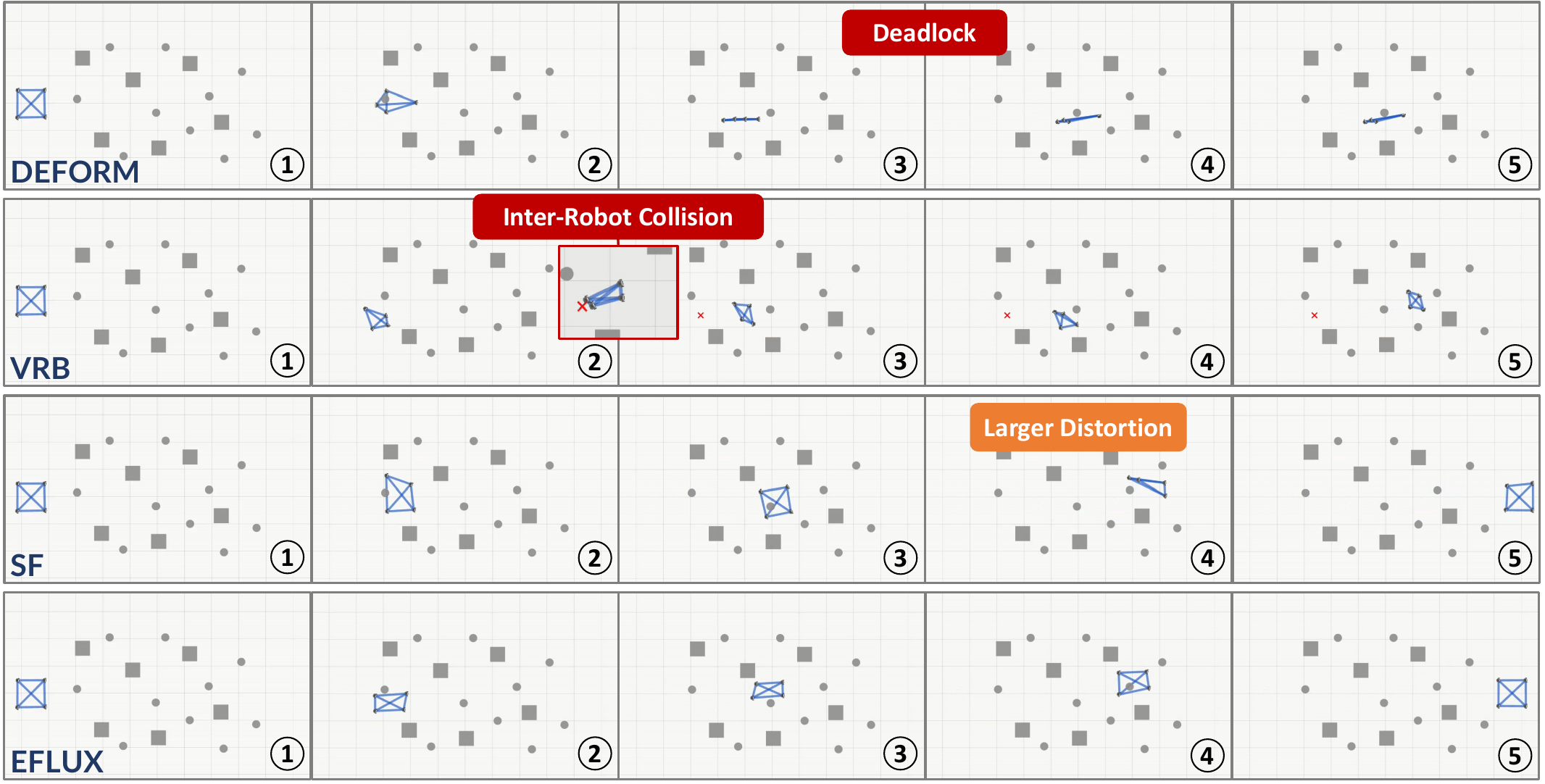}
      \label{fig:model_comparison_sparse}
    }
    \vspace{-0.2cm}
    \caption{Rows correspond to DEFORM, VRB, SF, and EFLUX. DEFORM becomes trapped in a deadlock, VRB experiences a deadlock or an inter-robot collision, and SF completes the passage but undergoes pronounced transient formation distortion. EFLUX maintains coordinated adaptation throughout traversal, reducing peak distortion and improving success rates.}
    \label{fig:approximation}
\end{figure}

The benchmark contains five constrained scenarios, summarized in Table~\ref{tab:main_comparison}. 
The first two scenarios involve narrow corridors with gap widths of \(1.2\,\mathrm{m}\) and \(0.7\,\mathrm{m}\) to evaluate formation adaptation under different passage widths. 
The third scenario contains a large obstacle wider than the nominal formation, requiring explicit split-merge reconfiguration. The final two scenarios use sparse and dense obstacle fields containing repeated local bottlenecks that require multiple formation adaptations during traversal.

\begin{figure*}[t]
\vspace{0.1cm}
      \centering
    \includegraphics[width=2\columnwidth]{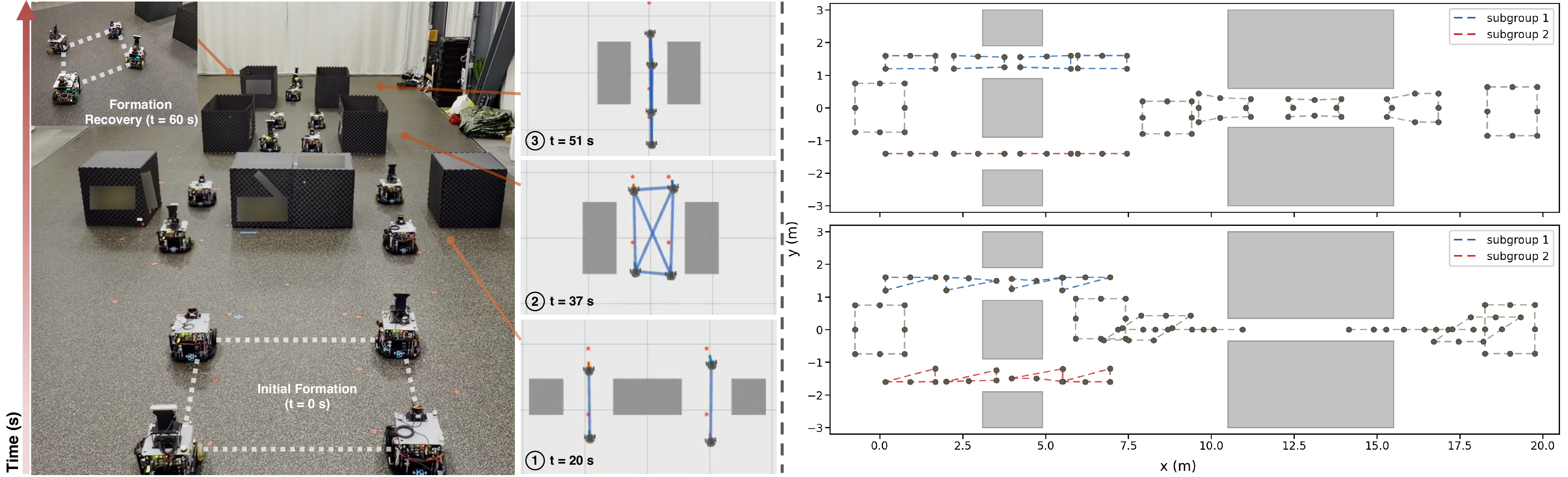}
    \vspace{-0.1cm}

    \caption{EFLUX in a challenging scenario requiring coupled deformation and reconfiguration. \textbf{Left}: real-robot execution on a three-phase map. The team starts from the nominal formation at $t=0$ s, splits into two subgroups to bypass central obstacles at $t=20$ s, merges and compresses at $t=37$ s, and collapses into a line formation at $t=51$ s before recovering the nominal formation at $t=60$ s. \textbf{Right}: two representative EFLUX Stage III plans for an 8-robot formation, both feasible yet visibly diverse. Black dots denote synchronized formation waypoints. In the reconfiguration phase, the plans split and merge around the obstacles in different ways; in the deformation phase, they traverse the $1.2$ m and $0.7$ m corridors by deforming into a near-nominal compressed formation and a line formation, respectively. These plans show the framework's adaptability to varying passage geometry.}

    \label{fig:challenging_scenario}
      \vspace{-0.4cm}
\end{figure*}

As shown in Table~\ref{tab:main_comparison}, EFLUX achieves strong performance across all environments and formation types. In the $1.2\,\mathrm{m}$ corridor, EFLUX achieves a $100\%$ SR with shorter paths than prior methods. In the more constrained $0.7\,\mathrm{m}$ corridor (Fig.~\ref{fig:model_comparison_corridor}), VRB fails due to local-minimum traps induced by potential-field-based obstacle avoidance at the entrance and inside the narrow corridor. DEFORM can switch to a line-like template based on feasible width, but template convergence can conflict with local obstacle avoidance and inter-robot safety constraints, causing deadlock during switching. SF traverses the corridor successfully but oscillates as each robot trades off obstacle avoidance against formation keeping inside the passage. In contrast, EFLUX uses high-level adaptation reasoning to plan waypoints that anticipate the bottleneck, attaining a $100\%$ SR with formation distortion comparable to the similarity-cost-optimized SF and shorter total paths.

The advantage of EFLUX becomes more evident in the large-obstacle and cluttered scenarios (e.g., Fig.~\ref{fig:model_comparison_sparse}). VRB lacks explicit topological adaptation, while DEFORM and SF remain constrained by template switching or multi-objective conflicts, which can lead to deadlock, collision, or pronounced transient formation distortion when split--merge behavior is required. By selecting when to preserve, deform, or split the team, EFLUX attains the highest or tied-highest SR in every configuration. Since $e_{\mathrm{sim}}$ is averaged only over successful runs, it inherently favors methods that fail in the harder cases; under this metric, EFLUX remains comparable to the similarity-cost-optimized SF. These results show that combining LLM-guided adaptation reasoning with deterministic geometric validation improves robustness and achieves balanced performance in constrained environments.

\subsection{Numerical Analysis of EFLUX}

We analyze EFLUX on challenging scenarios (Fig.~\ref{fig:challenging_scenario}) that contain multiple narrow regions and require both formation deformation and reconfiguration. 
This setting isolates the planning behavior of the agentic framework, which we analyze through module ablations, different LLM backbones, and varying team sizes to assess scalability.

\noindent \textbf{Stage-wise ablation study.}
We conduct stage-wise ablations to evaluate the contribution of each module in EFLUX. As shown in Table~\ref{tab:stage_ablation}, removing either Stage I or Stage II preserves the final success rate but reduces the one-shot success rate and increases the number of Stage III attempts. This indicates that semantic scene interpretation and explicit strategy reasoning mainly improve the quality of the initial waypoint proposal. Removing both stages further lowers one-shot success and raises first-attempt violations; direct waypoint generation is therefore less reliable without intermediate scene and strategy representations. The retry mechanism has the largest effect on robustness; with a single open-loop Stage~III attempt, the SR drops to $50.0\%$, showing that feedback-guided retry is critical for recovering from invalid waypoint proposals. Removing waypoint refinement also degrades performance and increases planning time, token usage, and replanning attempts, suggesting that local refinement helps correct minor geometric violations before triggering more costly retry rounds.

\begin{table}[th]
  \centering
  \caption{Stage-wise ablation study.}
  \label{tab:stage_ablation}
  \scriptsize
  \setlength{\tabcolsep}{2.0pt}
  \renewcommand{\arraystretch}{1.06}
  \begin{threeparttable}
  \begin{tabular*}{\columnwidth}{@{\extracolsep{\fill}}lccccc@{}}
    \toprule
    Variant
    & SR\tnote{\dag}\,\upmetric
    & $T_{\mathrm{plan}}$ (s)\,\downmetric
    & $N_{\mathrm{att}}$\,\downmetric
    & $V_{\mathrm{1st}}$\,\downmetric
    & $N_{\mathrm{tok}}$\,\downmetric \\
    \midrule

    \rowcolor{gray!12}
    Full pipeline
    & 100.0\% & 429.88 & 1.15 & 0.60 & 97.1k \\

    \midrule
    w/o Stage~I
    & 100.0\% & 430.32 & 1.50 & 1.50 & 88.6k \\

    w/o Stage~II
    & 100.0\% & 358.39 & 1.50 & 0.95 & 90.6k \\

    w/o Stage~I \& II
    & 95.0\% & 266.72 & 1.90 & 2.30 & 75.5k \\

    w/o retry mechanism, open loop
    & 50.0\% & 414.36 & 1.00\tnote{\ddag} & 1.20 & 89.2k \\

    w/o waypoint refinement
    & 95.0\% & 477.36 & 1.58 & 1.53 & 111.4k \\

     w/o geometric grounding 
    &60.0\%  &526.12  &2.42  &3.58  &126.3k  \\

    \bottomrule
  \end{tabular*}

  \begin{tablenotes}[flushleft]
    \scriptsize
    \item Each variant uses 20 trials. [\dag] Retry budget of $5$ attempts; a run is marked failed once the budget is exhausted. [\ddag] No retry by design.
  \end{tablenotes}
  \end{threeparttable}
\end{table}

\begin{figure}[th]
      \centering
    \includegraphics[width=0.99\columnwidth]{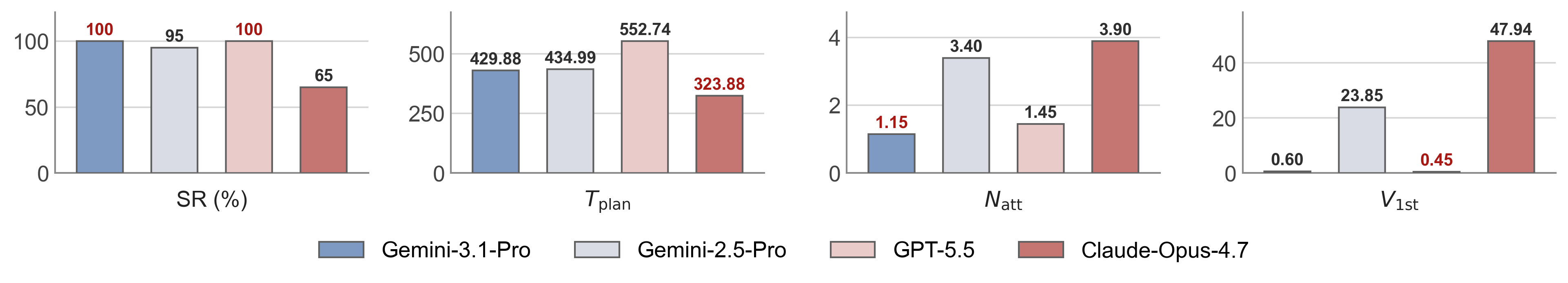}
    \vspace{-0.3cm}
      \caption{Performance comparison using different LLM backbones. All backbones use 20 trials on the same map.
      }
      \label{fig:model_comparison}
\end{figure}

\noindent \textbf{LLM backbone comparison.}
In Fig.~\ref{fig:model_comparison}, we evaluate the same planning pipeline with different LLM backbones, including Gemini~\cite{deepmind2026gemini3}, OpenAI models~\cite{OpenAIAPI2026}, and Claude~\cite{AnthropicClaudeAPI2026}. For Gemini, we additionally compare two model variants to assess the sensitivity of EFLUX to model reasoning capability. All models are evaluated on the same map with identical planner settings. This experiment examines whether the proposed framework is tied to a specific LLM backbone or can generalize across different foundation-model families.

\noindent\textbf{Scalability with respect to robot number.}
We evaluate scalability by fixing the square formation footprint at $1.5\,\mathrm{m} \times 1.5\,\mathrm{m}$ and increasing the team size from $4$ to $6$ and $8$ robots. 
As shown in Table~\ref{tab:scalability_robot_number}, EFLUX achieves a $100\%$ SR for both the $4$- and $6$-robot teams. As the team size increases within the same formation footprint, planning time, token usage, and first-attempt blocking violations all increase, since more robots must be coordinated under higher spatial density. For the $8$-robot team, the SR decreases to $60\%$, indicating that valid waypoint generation becomes more challenging under the fixed retry budget. Fig.~\ref{fig:challenging_scenario} (right) shows representative successful 8-robot plans passing different sizes of the narrow corridors.

\begin{table}[!th]
  \centering
  \caption{Scalability w.r.t.\ robot number.}
  \label{tab:scalability_robot_number}
  \scriptsize
  \setlength{\tabcolsep}{2.0pt}
  \renewcommand{\arraystretch}{1.06}
  \begin{threeparttable}
  \begin{tabular*}{\columnwidth}{@{\extracolsep{\fill}}lccccc@{}}
    \toprule
    Robots
    & SR\,\upmetric
    & $T_{\mathrm{plan}}$ (s)\,\downmetric
    & $N_{\mathrm{att}}$\,\downmetric
    & $V_{\mathrm{1st}}$\,\downmetric
    & $N_{\mathrm{tok}}$\,\downmetric \\
    \midrule
    $N=4$ & 100.0\% & 429.88 & 1.15 & 0.60  & 97.1k  \\
    $N=6$ & 100.0\% & 536.79 & 1.65 & 4.80  & 123.1k \\
    $N=8$ & 60.0\%  & 581.13 & 1.75 & 10.00 & 139.0k \\
    \bottomrule
  \end{tabular*}

  \begin{tablenotes}[flushleft]
    \scriptsize
    \item[] All runs use the Gemini-3.1-Pro backbone; each team size uses 20 trials.
  \end{tablenotes}
  \end{threeparttable}
\end{table}

\subsection{Hardware Experiments}

We further validate the proposed framework through indoor hardware experiments using multiple mobile ground robots.
The robots perform collaborative tasks in cluttered environments while adaptively maintaining formation. Fig. \ref{fig:challenging_scenario} (left) shows representative snapshots illustrating coordinated and collision-free formation adaptation in real-time.

\section{Conclusion}

We propose \Method, an elastic multi-robot formation navigation framework to reason over continuous formation deformation and topology-changing reconfiguration. 
By integrating deterministic geometric grounding with LLM-guided adaptation planning, \Method enables coordinated and robust navigation through complex constrained environments. 
Results from simulation and hardware experiments demonstrate improved success rates and reduced deadlock compared with representative state-of-the-art baselines. 
Future work will extend the framework to dynamic environments, heterogeneous teams, and long-horizon open-world coordination.

\bibliography{references}
\end{document}